\documentclass[conference, 12pt]{IEEEtran}
\usepackage{cite}
\usepackage{amsmath,amssymb,amsfonts}
\usepackage{algorithmic}
\usepackage{tabularray}
\usepackage{graphicx}
\usepackage{textcomp}
\usepackage{fancyhdr}
\usepackage{xcolor}
\usepackage{float}
\usepackage{booktabs}
\usepackage[T1]{fontenc}
\def\BibTeX{{\rm B\kern-.05em{\sc i\kern-.025em b}\kern-.08em
    T\kern-.1667em\lower.7ex\hbox{E}\kern-.125emX}}
\usepackage{fancyhdr}
\usepackage{url}
\usepackage{graphicx}
\usepackage[normalem]{ulem}
\usepackage{subcaption}
\usepackage[font=bf]{caption}
\usepackage{float}
\usepackage{xspace}
\usepackage{enumitem,kantlipsum}
\usepackage{rotating}
\usepackage[hidelinks]{hyperref}
\usepackage{dotlessi}

\newcommand{\dotsaag}[0]{\raisebox{-1ex}{\includegraphics[height=0.4ex]{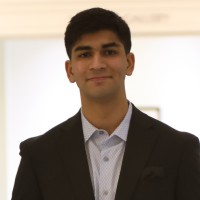}}}
\fancypagestyle{IEEEtitlepagestyle}{
  \fancyhf{} 
   
  \setlength{\headheight}{15.31181pt}
  \fancyhead[R]{\raisebox{-1ex}{Duke Robotics Club | \thepage}} 
}
\begin{document}

\title{Technical Design Review of Duke Robotics Club's Oogway \& Crush: AUV's for RoboSub 2026}

\author{Patrick Zheng, Saagar Arya, Hung Le, Mathew Chu, Nathanael Ren, Niko Weaver,\\ Isabella Chen, Jill Wang, Raine Cheng, Siddharth Kini, Avrick Altmann, Srinath Iyer,\\ Ivan Chen, Ian Suh, Parker Jones, Pierson Jones, Sebastian deSouza,\\ Suhaani Sriram, Suvas Aggarwal}

\maketitle

\begin{abstract}
\normalsize
The Duke Robotics Club presents Oogway and Crush, our AUVs for RoboSub 2026. This year's strategy expands on our previously narrowed scope, targeting all four of RoboSub's design goals for the first time: movement, vision, manipulation, and acoustic tracking. This expansion is based on sustained reliability investment across all three subsystems. Mechanically, Crush gained two additional thrusters and a CFD-optimized case, providing pitch stability. Electrically, we addressed accumulated failure points by repairing unreliable connections and upgraded thruster control hardware. We also redesigned our acoustics system, adding a new custom PCB with higher-order filters, significantly improving pinger detection reliability. On the software side, improvements to state estimation, sonar-based object detection, vision-driven task planning, and IVC enable more capable and coordinated autonomous runs. Paired with investments in our testing infrastructure to maximize our limited pool time, we can now attempt a broader set of tasks while maintaining the reliability our competition strategy demands.

\end{abstract}

\section{Competition Strategy}

Our strategic vision centers on \textit{core tasks}, which our robots are programmed to perform every run, and \textit{alternate tasks} which may be attempted at the end of a successful run for additional points. \textit{Core tasks} are implemented with fail-safes to ensure reliability and are delegated between robots to maximize efficiency. With limited pool testing time, every new task competes directly against reliability testing for existing ones, driving our core/alternate distinction and task prioritization. This year's sustained investment across mechanical, electrical, and software systems gives us the confidence to expand our scope to all four of RoboSub's design goals for the first time. The tradeoffs between complexity and reliability for each task are discussed below.

\begin{figure}[H]
     \centering
     \begin{subfigure}[b]{0.45\textwidth}
         \centering
         \includegraphics[width=0.9\textwidth]{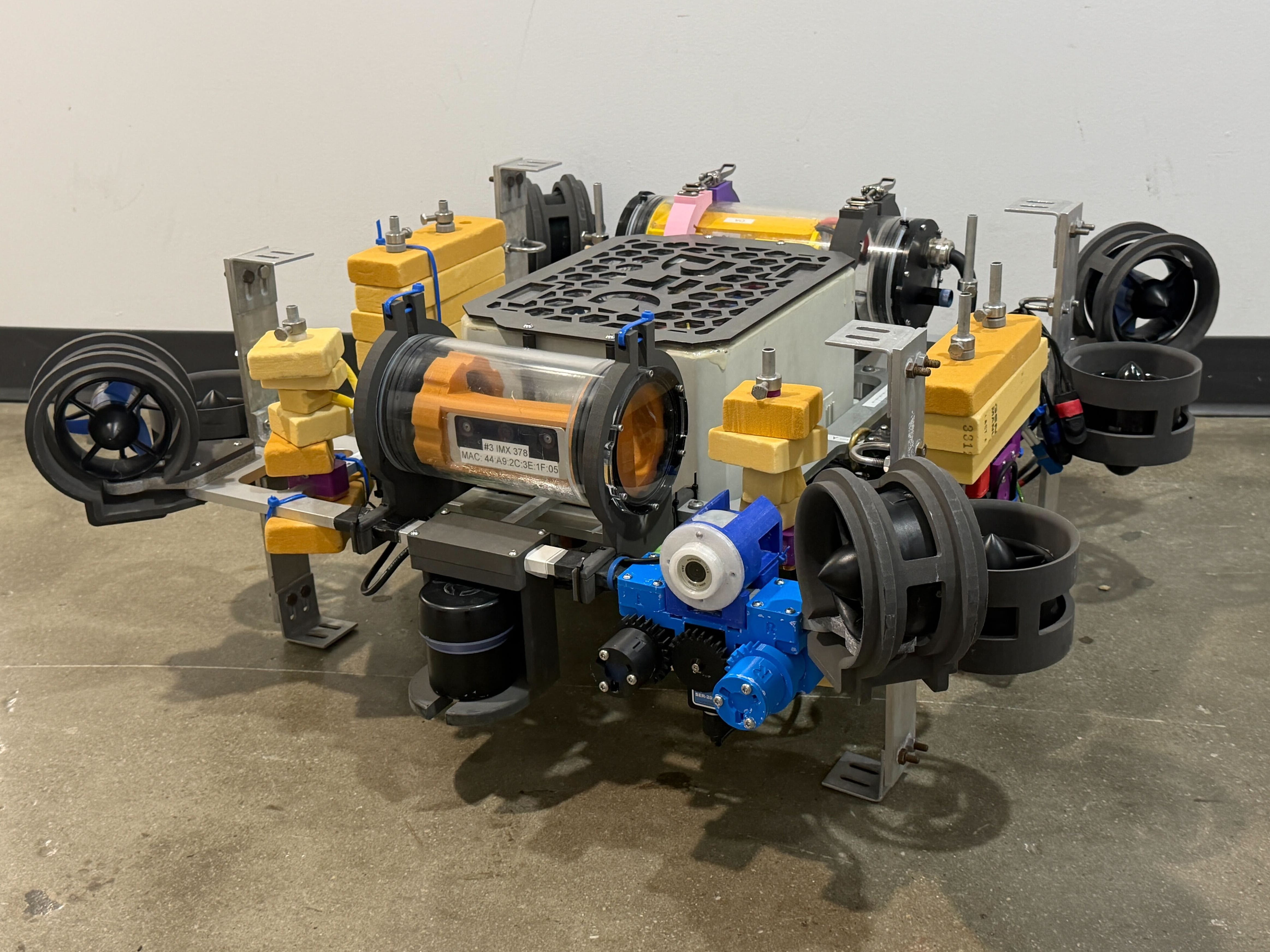}
         \caption{Oogway}
     \end{subfigure}
     \hfill
     \begin{subfigure}[b]{0.45\textwidth}
         \centering
         \includegraphics[width=0.9\textwidth]{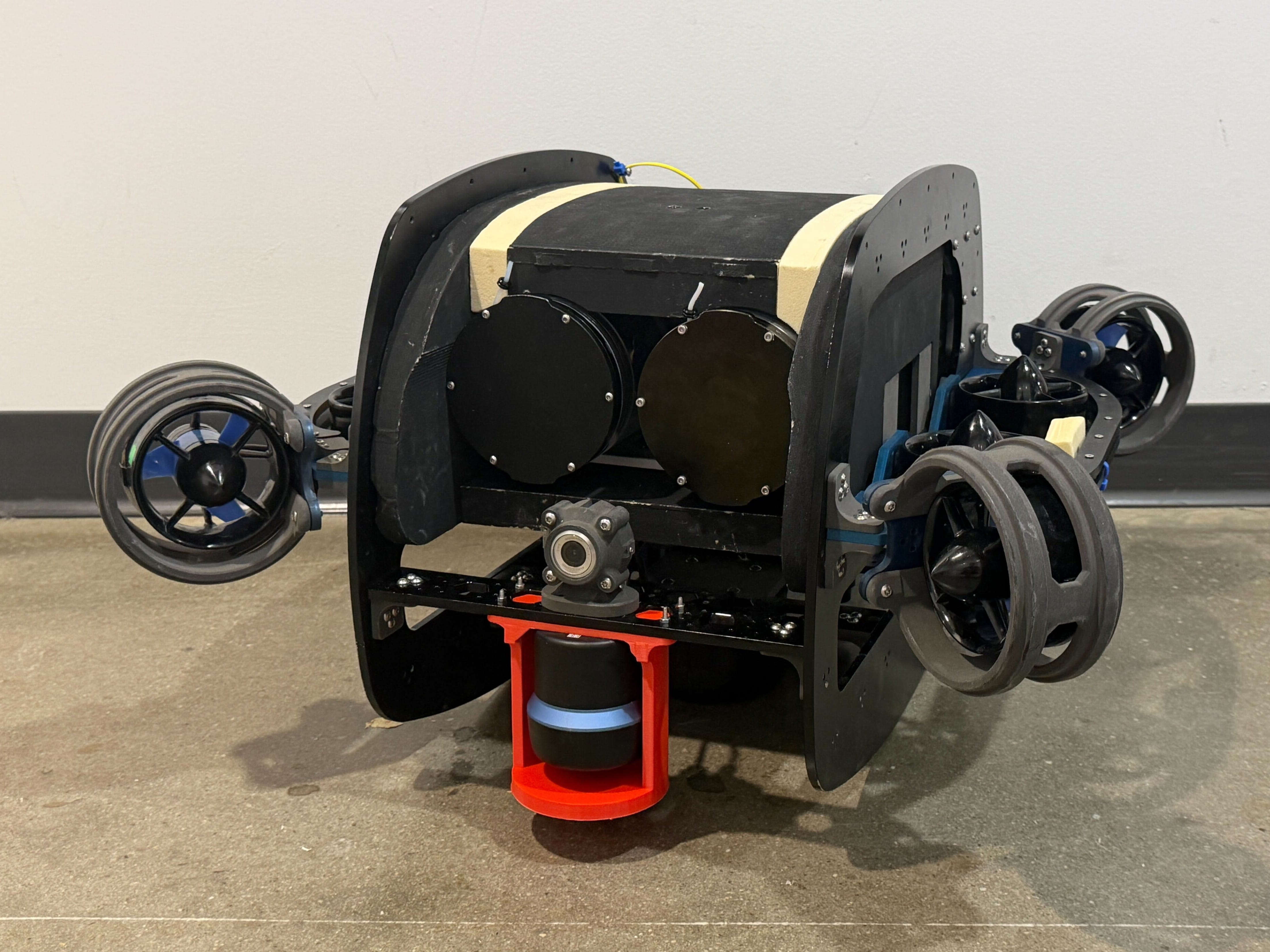}
         \caption{Crush}
     \end{subfigure}
     \caption{Duke Robotics Club's AUVs for RoboSub 2026}
\end{figure}

\subsection{Heading Out -- Gate}
Both robots must complete the gate task each run. To maximize reliability, we employ a multimodal approach: our CV subsystem detects the gate from arbitrary angles and estimates symbol positions. As a fallback, our control system can perform dead reckoning through the gate using odometry and state estimates. Both robots attempt the coin flip bonus; Crush performs the barrel roll style task while Oogway skips it to preserve state.

\subsection{Avoid Debris -- Slalom}
Navigating the channel is a \textit{core task} for Crush. After completing the gate task, Crush dead reckons through the three sets of slalom poles. We experimented with HSV filtering to locate each pole, but the various positions of the 9 poles and case logic proved difficult to reliably implement, and dead reckoning with our improved movement stack proved more consistent and reduced complexity. Oogway will not attempt the slalom, reserving time to search for and navigate to the torpedo banner.

\subsection{Deploy -- Torpedoes}
The torpedo task is a \textit{core task} for Oogway. After the gate, Oogway uses its front-facing camera and YOLO model to navigate toward the torpedo board. Once positioned, it switches to sonar for yaw alignment and HSV filtering to center on the opening. Sonar alignment was chosen over vision for robustness to the lighting variability and turbidity common in competition pools.

Before firing, Oogway queries its hydrophone array to check whether the random pinger is nearby. If so, Oogway fires first, then signals Crush to surface in the octagon. Otherwise, Oogway signals Crush first, waits for acknowledgment of octagon completion, then fires. This sequencing enables reliable coordination for the random pinger bonus. Timeouts at each IVC checkpoint ensure neither robot stalls on a missed message.

\subsection{Resupply -- Octagon}
The octagon is a \textit{core task} for Crush for the first time this year, made possible by improvements to our navigation stack. After the slalom, Crush uses sonar sweeps to locate the octagon's back wall and navigate inside. Once inside, Crush uses its downward-facing camera to locate the table, positions itself above it, scans the hanging images, and rotates to face the correct one. Crush then waits for an IVC message from Oogway before surfacing. As our grabbing mechanism is still in development, we will not attempt the object manipulation portion of this task.

\subsection{Recon -- Bins \& Return Home}
Both tasks are designated \textit{alternate tasks}. Bin navigation currently requires an extensive search over a large area, making it too time-inefficient to run as a core task, and the engineering investment to improve it was better spent hardening the torpedo and octagon pipelines. Both robots carry marker droppers and may attempt bins opportunistically at the end of a run. Return to gate requires global position estimates we do not have at competition fidelity, and offers relatively few points compared to the \textit{core tasks}. 
\section{Design Strategy}

\subsection{Mechanical Design}

Mechanical work this year focused on reliability improvements to Oogway and expanding Crush's capabilities to support our broader competition scope.

\subsubsection{Reliability}
During transport and testing, critical connections have come loose or been damaged; a momentary battery disconnect shuts the robot down entirely, requiring it to be pulled from the pool and restarted, wasting up to 5 minutes of pool time. To address this, Oogway's epoxied connections were reinforced and battery connectors were replaced with hard-soldered and potted alternatives. Another example is our torpedo launcher's large profile which made it prone to snapping when passing through doorways, frequently preventing in-water torpedo testing. To fix this, we reduced the total volume by more than 15\%. We receded the launch tubes further into the robot, and used structural screws and inset nuts to strengthen the weakest parts that previously snapped.

\subsubsection{Crush Eight Thrusters}
Two additional vertical thrusters were added by redesigning the side wings and mounting system (Fig. 2), providing the pitch control needed to reliably navigate to and surface inside the octagon.
\begin{figure}[h]
    \centering
    \includegraphics[width=0.7\linewidth]{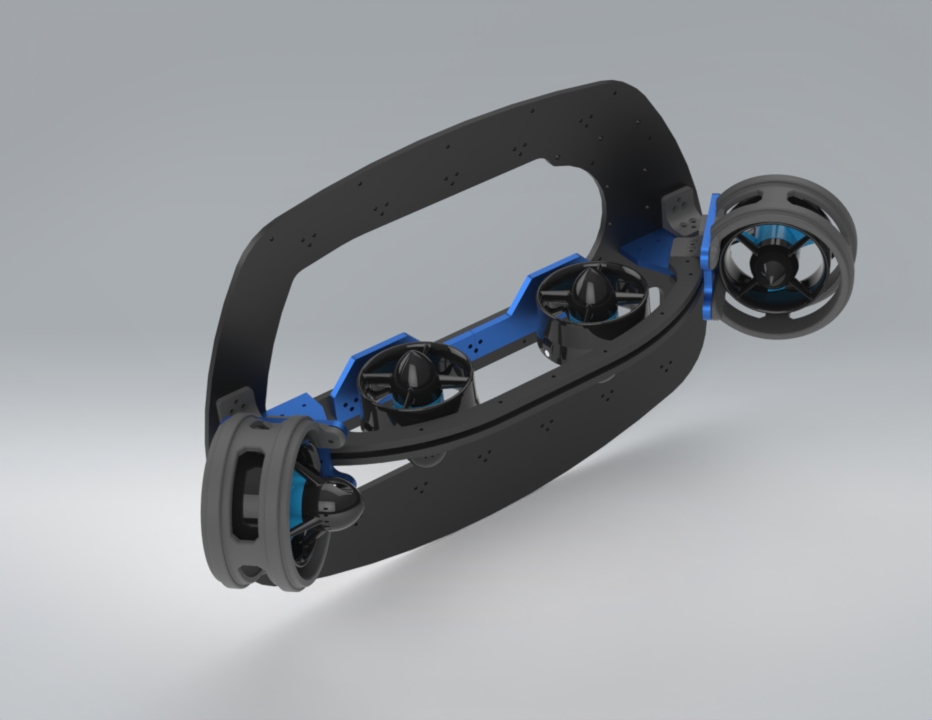}
    \caption{Side Thruster Configuration}
    \label{fig:thrusters}
\end{figure}

\subsubsection{Case CFD and Hydrodynamic Drag}
The case was designed to reduce drag on the robot. We evaluated the case design via CFD in Ansys Fluent. To extract the total drag force ($F_D$), the solver numerically integrated the pressure and viscous shear stresses over the AUV's surface area ($S$) using $F_D = \int_S (p \mathbf{n} \cdot \mathbf{i} + \mathbf{\tau}_w \cdot \mathbf{i}) dA$, where $p$ is static pressure, $\mathbf{\tau}_w$ is wall shear stress, $\mathbf{n}$ is the surface normal, and $\mathbf{i}$ is the flow direction vector. This surface integration verified a $29\% \pm 16\%$ total drag reduction over the bare chassis. The CFD simulations were visualized to show the water flow (Fig. 3).

\begin{figure}[h]
    \centering
    \includegraphics[width=0.5\textwidth]{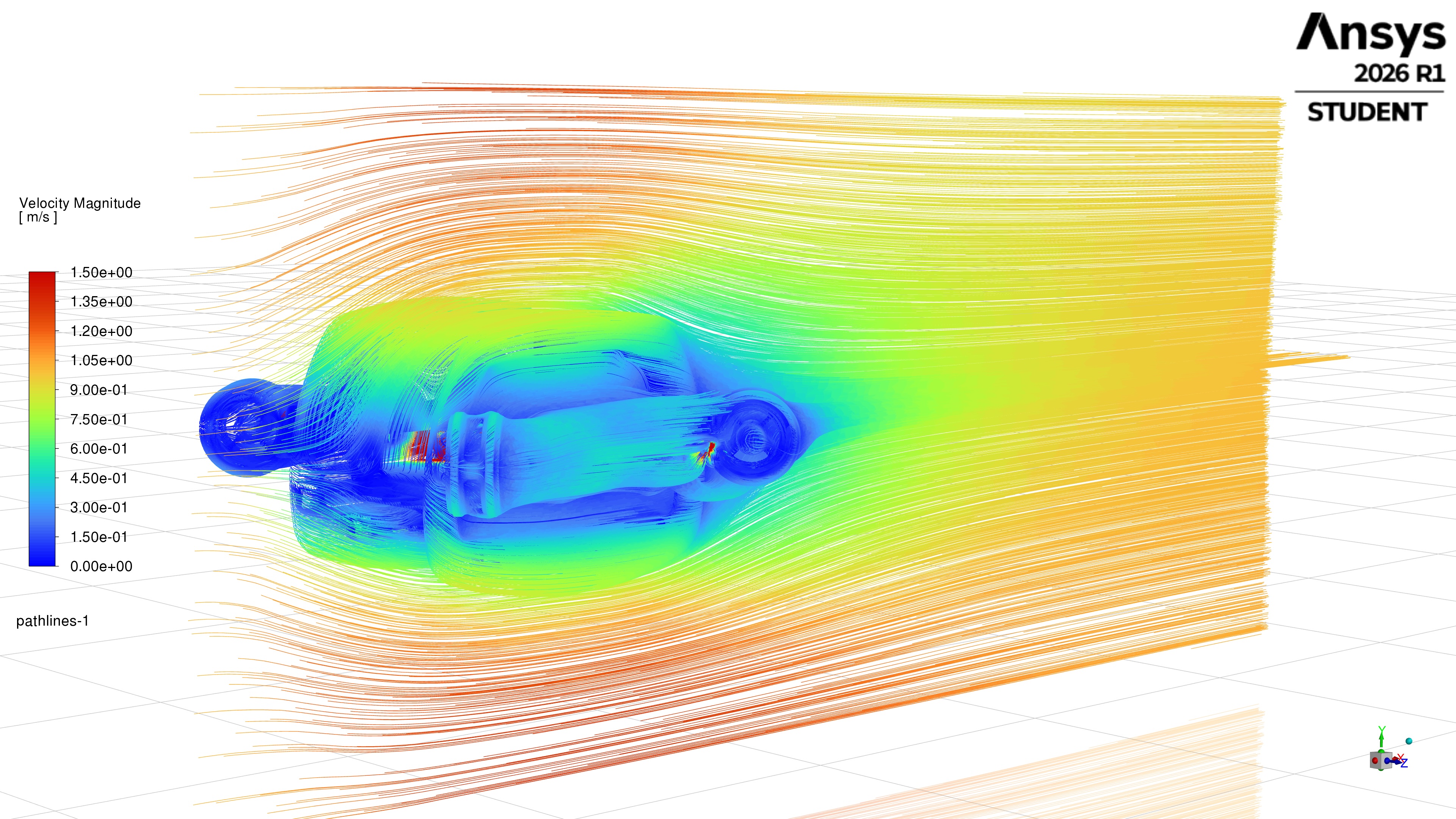}
    \caption{CFD Plot of velocity flow lines}
\end{figure}

\subsubsection{Buoyancy}
Crush's previous buoyancy foam consisted of flat sheets mounted externally, creating turbulent flow that degraded movement consistency. We custom CNC-milled buoyancy blocks to match the inner profile of the case,contributing no additional drag while maintaining positive buoyancy (Fig. 4). The blocks were designed to be removable in under 1 minute to not impede capsule access during critical troubleshooting.
\begin{figure}[H]
    \centering
    \includegraphics[width=0.45\linewidth]{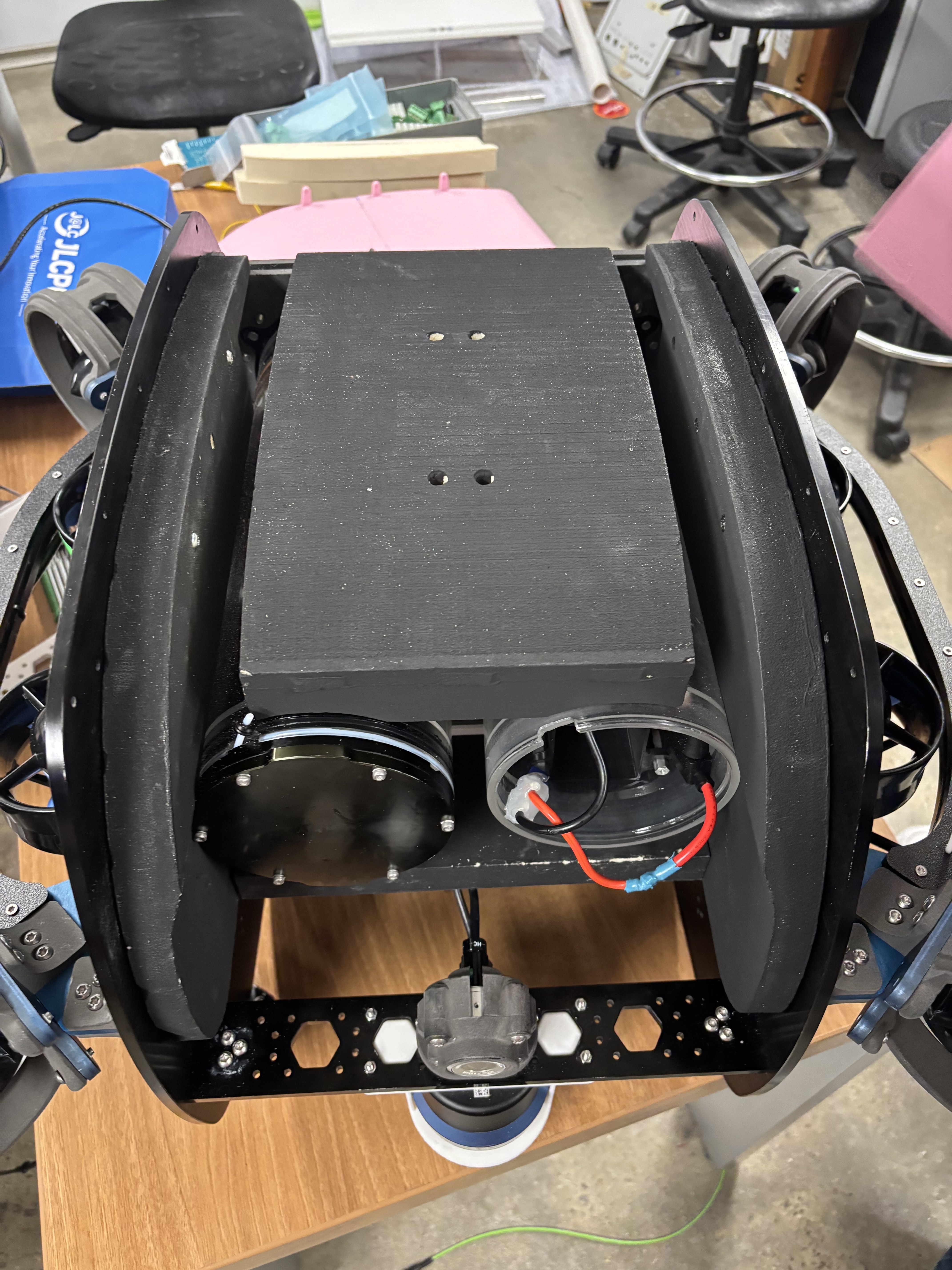}
    \caption{Custom Buoyancy Blocks}
    \label{fig:buoyancy}
\end{figure}

\subsubsection{Marker Dropper}
With increased confidence in Crush's movement capabilities and for additional flexibility in completing the bins task, we decided to redesign and install a marker dropper on Crush. We used the same 19 mm brass spheres, but now loaded into a capped cylindrical barrel with a paddle-like servo attachment controlling each drop.

\begin{figure}[H]
    \centering
    \includegraphics[width=0.8\linewidth]{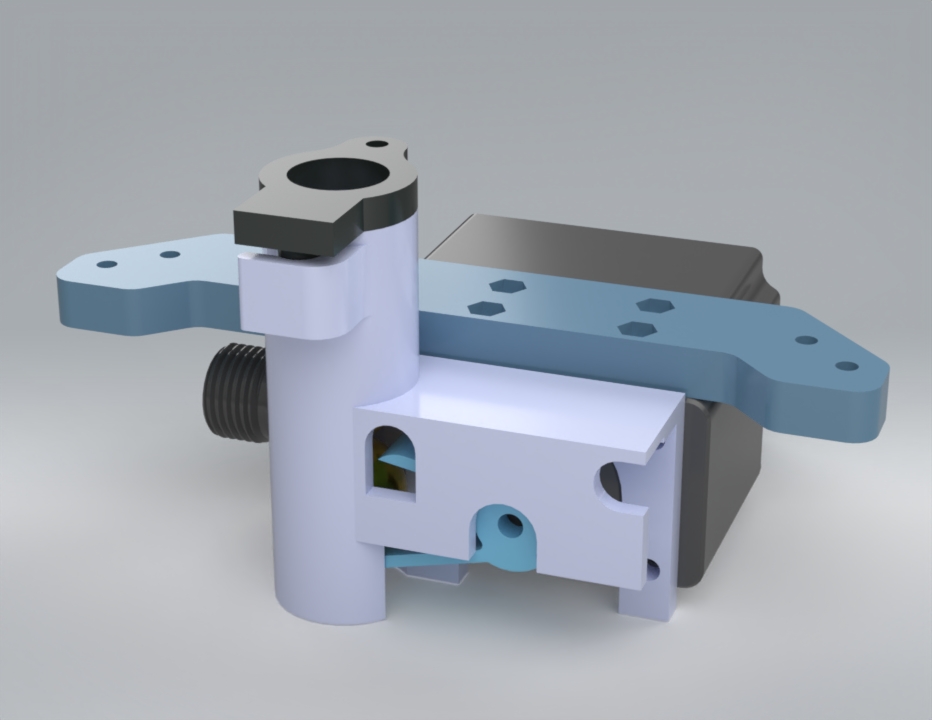}
    \caption{Crush Marker Dropper}
    \label{figSSmarker_dropper}
\end{figure}
\newpage
\subsection{Electrical Design}
Many of our persistent electrical issues stemmed from compounding small failures (loose connections, inconsistent signal routing, and damaged wires) and unnecessary complexity (additional Ethernet adapters, multiple Arduinos with empty ports, etc.). A single unreliable connection would cripple an entire subsystem during a pool test, resulting in wasted testing and engineering time diagnosing the issue in our overly complex stack. The team's goal this year was to strengthen the quality of and simplify our stack where possible.

Post-competition, we performed a detailed inspection of both robots and repaired unreliable connections, exposing the most vulnerable parts of the current architecture. This informed our planned PCB-based stack redesign for 2027 (See Appendix B).

\subsubsection{Acoustics}
Our acoustics system features a custom PCB with four 8th-order Butterworth bandpass filters followed by per-channel amplification. The higher-order filter provides a steep frequency rolloff that effectively isolates the pinger signal from thruster noise. As shown in Fig. 6, the filtered signal's amplitude swings $\pm$2V and the frequency domain analysis confirms energy is concentrated within the filter passband, yielding a high signal to noise ratio for reliable downstream detection.

\begin{figure}[H]
    \centering
    \includegraphics[width=1.0\linewidth]{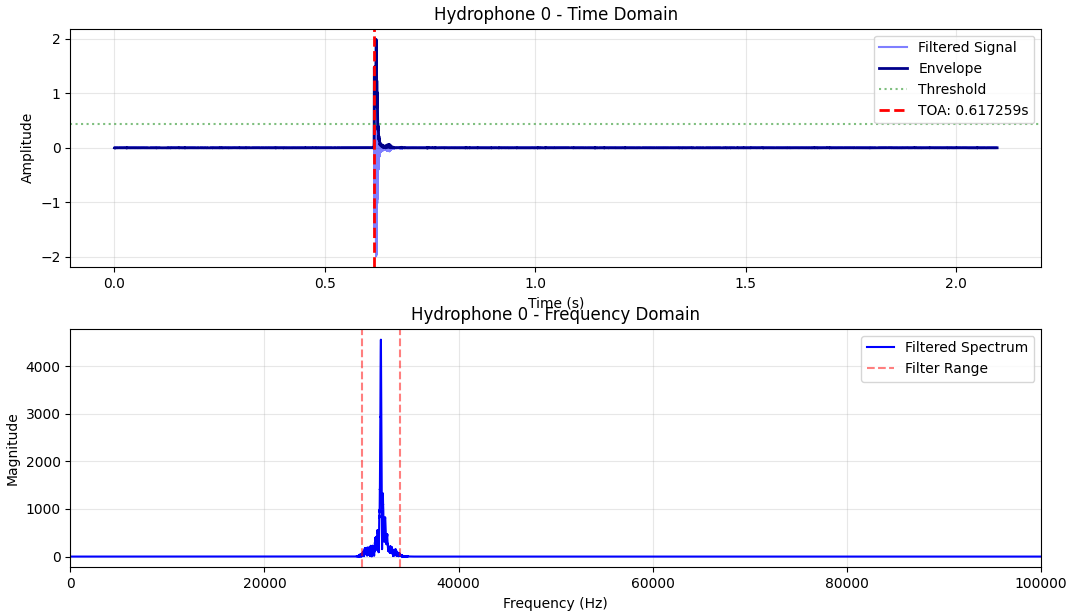}
    \caption{Filtered hydrophone signal time domain (top) and frequency spectrum (bottom)}
\end{figure}

On the software side, three analyzers are run on each recording. A validity analyzer confirms the snippet is complete and contains a ping. A Time-of-Arrival (TOA) analyzer identifies which hydrophone registers first, giving quadrant-level bearing. A Nearby analyzer uses feature extraction on the raw and filtered signals with a Random Forest classifier to determine whether the pinger is within 10 feet, achieving 93.96\% $\pm$ 1.35\% accuracy across 5-fold cross-validation on 447 samples.

\subsubsection{Inter-Vehicle Communication (IVC)}
Last year's Waterlinked M-16 modems enabled IVC but suffered frequent disconnects, resulting in unreliable completion of the IVC task. This year, modems are directly connected to a USB port on the central computer and powered by dedicated voltage regulators, reducing intermittent dropouts and disconnects.

\subsection{Software Design}
This year, the software team focused on improving our ROS 2 autonomy stack, emphasizing state-estimation reliability, perception-driven task planning, IVC, and poolside debugging. Our robot-agnostic architecture allows Oogway and Crush to reuse the same core packages while maintaining robot-specific configurations, letting us build and test new features across both robots without duplicating large portions of the codebase.

\subsubsection{State}
Accurate state is essential for every autonomous behavior, but we encountered persistent issues where depth, orientation, or fused state information could become inconsistent across topics. To address this, we made the pressure sensor the universal source of depth, so that the z-position reported in \texttt{/state} directly matches the depth reported by \texttt{/sensors/depth}. This gives controls and task planning a single trusted depth estimate. In parallel, we worked with vendors to debug sensor behavior and driver issues, separating software-side filtering problems from device-level ones.

\subsubsection{Sonar Object Detection}
Previously, sonar was primarily used as a low-level sensor stream, but this year we worked toward extracting higher-level information, such as the location and normal angle of objects. This allows sonar to serve as a perception source for tasks where camera-based detection is unreliable. By converting sonar returns into object and wall estimates, the task planner can use sonar as a source of relative object position that can drive navigation. Having this backup system increases the situations in which our robot can confidently operate in.

\subsubsection{Acoustics Integration}
Acoustics were integrated into the ROS 2 stack through a dedicated acoustics service that wraps hydrophone-processing code and exposes results through ROS topics and service calls. The outputs include whether the acoustics system is ready, whether the pinger is getting closer, and the estimated direction of the pinger. Task planning can then request or subscribe to this information without depending on the internal details of signal processing, keeping the two layers independently maintainable. This structure also supports IVC integration: once one robot has useful pinger information, it can communicate that to the other robot to help determine task order.

\subsubsection{Task Planning}
In task planning, we focused on making vision-based movement more responsive. We rewrote the yaw-to-CV-object behavior so that the robot updates its yaw setpoint continuously from the detected object angle instead of rotating to a fixed angle, stopping, and checking again. This makes the robot’s alignment behavior faster and smoother, especially when the object is moving in the camera frame. We also began extending this approach from simply yawing toward a detected object to moving toward CV objects, allowing task-planning logic to use visual detections more directly when generating motion goals. This work improves tasks that require precise alignment.

Since both Oogway and Crush run tasks during the same competition attempt, IVC must be reliable enough to support real-time coordination rather than simply logging messages after the fact. This year, we restructured the IVC task-planning code to remove asynchronous file I/O issues, simplify the relationship between communication utilities and competition tasks, and create generic functions usable by both robots. The goal is for Oogway and Crush to share the same IVC task logic while still allowing each robot to make decisions based on its assigned role, current task, and information received from the other robot. This makes our runs more efficient and maximizes points for IVC.

\subsubsection{Foxglove GUI}
Pool time is our scarcest resource, and as Oogway gets older we have increasingly wasted pool time debugging issues rather than testing. To address this, we refined our Foxglove-based GUI with improved sensor status panels that correctly detect when sensors stop publishing, filter displayed sensors, and share common parsing logic across panels, making it easier to verify both robots are publishing expected data before beginning an autonomous run. We also added a keyboard-control panel mapping six degrees of freedom to keyboard inputs, removing the physical joystick dependency for motion testing. Finally, we set up Foxglove Agent support on both robots so that recorded data can be automatically imported from onboard bag files, improving our ability to review pool tests.

\section{Testing Strategy}
Our testing follows a staged approach designed to preserve pool time: components are validated independently before integration, simulation and bench testing are front-loaded to catch issues early, and pool sessions are reserved for integration testing and tasks that require physical responses and real-world conditions. This mirrors our broader philosophy that pool time is our scarcest resource.

\subsection{Simulated Testing}
\subsubsection{Bag Files}
Bag files are our primary tool for testing software outside of the pool. By replaying recorded sensor data and code output, we can both diagnose unexpected behavior from past runs and validate new code against real-world data without requiring pool time. This year we implemented system and software-level optimizations to ensure bag files can be collected reliably in almost all situations, directly addressing gaps from last season. Bag files provide two key debugging capabilities: (1) seeing the precise outputs of our code and how it interprets real-world environments to determine what led to unexpected behavior, and (2) testing new code on real-world sensor data. Together these allow us to front-load as much software validation as possible, reserving pool time for tests that require physical responses and movement.

\subsubsection{Acoustics}
For competition, the Nearby analyzer is the primary focus. Three recordings are collected, taking 15 seconds total, and a majority vote is taken. Since the classifier is 94\% accurate, the probability of it being wrong on at least two of three independent recordings is 1.04\%, giving significantly higher confidence than any single sample. Across two semi-final runs and one finals run, the probability of all measurements being correct is $>$96.9\%. The TOA analyzer is functional but reserved for future seasons as the primary method for navigating to tasks in low-visibility environments.

\subsection{Empirical Testing}

\subsubsection{Mechanical}
All water-facing components underwent leak testing before pool integration. Oogway's battery cable was bonded directly into the robot to eliminate a persistent water intake failure point. Crush required the same leak-check process after major wiring work to support eight thrusters. Crush's new buoyancy blocks were validated against CAD measurements, with close agreement between predicted and actual center of mass and buoyancy, requiring only minor pool-side adjustments. This sim-to-real fidelity meant Crush behaved as expected from the outset. Crush's pitch behavior at speed was tested by pushing the robot forward and monitoring for deviations; after tuning, the robot moved straight with no discernible deviation, improving odometry accuracy at higher speeds. 

\subsubsection{Electrical}
Each electrical component was validated independently before stack integration. Onboard computers were run for 4-hour intervals under standard CV loads to verify sustained performance. Sensor outputs were recorded and compared to expected values. The thruster control path (Arduino Nano $\to$ PCA9685 $\to$ BlueRobotics ESC) had its output signal measured at each stage to check for skew, noise, and distortion, and thrusters were subsequently tested at various speeds for several 5-minute intervals. Modem reliability was tested by transmitting data packets across multiple trials to confirm consistent delivery. All components received a final function confirmation before pool testing began. Characterization data such as startup current and voltage draw were recorded to better understand nominal behavior, which informed and accelerated debugging efforts.

\subsubsection{Software}
Where possible, software systems were validated on land or in small water tubs before pool testing. IVC, for example, was first validated in a small bin, confirming messages were sent and received correctly and that each robot executed the expected behavior upon receipt, before moving to the pool to verify end-to-end coordination under real conditions. Before each pool run, sensor health is verified through our Foxglove GUI, confirming all expected topics are publishing before beginning an autonomous run. PID gains and state estimation are tuned iteratively across pool sessions, with bag files recorded and reviewed after each session to diagnose issues and validate fixes before returning to the pool.

\subsection{Bug Tracking and Integration Testing}
Pool testers file a standardized bug report form for every electrical and mechanical issue observed during pool sessions. The software team tracks bugs through GitHub Issues. For systems under active monitoring -- such as modem disconnects -- a report is filed every time the issue occurs, allowing us to quantitatively track whether fixes are reducing failure rates over time. This structured approach lets us distinguish one-off failures from systemic issues and prioritize fixes accordingly. 

\section{Acknowledgments}
The Duke Robotics Club is primarily sponsored by Duke University's Pratt School of Engineering. 

We are grateful to Ali Stocks and the Foundry staff for housing and supporting us. We are also indebted to Director of Undergraduate Student Affairs, Tarina Argese, for helping us plan events and our advisor, Professor Boyuan Chen, for technical assistance, invaluable club advice, and support. We also want to specifically thank Dean Jerome Lynch of the Pratt School for his continued support of our program. 

Furthermore, we owe our success to our sponsors: General Motors, the Duke Engineering Alumni Council, the Duke Student Organization Finance Committee, and Hudson River Trading, whose generous financial contributions made this year's competition effort possible. 

Finally, we thank RoboNation, whose commitment to RoboSub empowers student engineers like ourselves to explore our passion for robotics.

\clearpage
\appendices
\section{Mechanical Upgrades}
\subsection{Eight Thrusters}
Many iterations of the new thruster mounting were considered and tested using 3D printed prototypes. We had to take into account weight and the allowable positions of each thruster to not conflict with the diagonal thrusters, in addition to spacing the vertical thrusters far enough to have a meaningful impact on pitch control. New HDPE thruster guards connected the two diagonal thruster brackets, which are both flexible and lightweight. Overall, the addition of two more thrusters didn't increase the weight much due to the transition from the large aluminum side wing to smaller brackets.

\begin{figure}[h]
    \centering
    \includegraphics[width=0.6\linewidth]{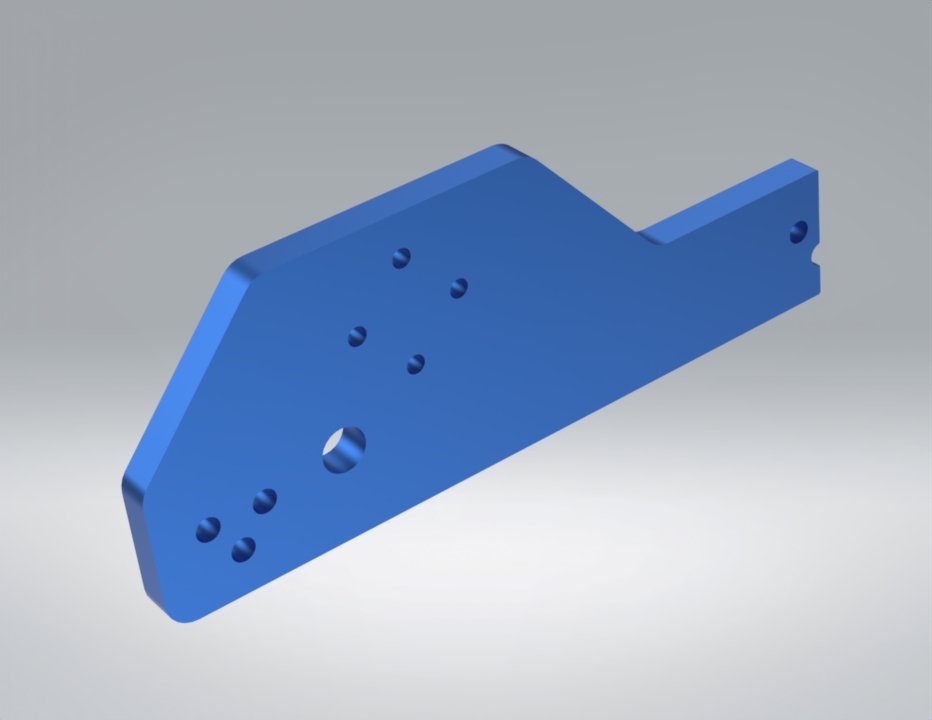}
    \caption{Vertical Thruster Bracket}
    \label{fig:vertical_bracket}
\end{figure}

\begin{figure}[h]
    \centering
    \includegraphics[width=0.6\linewidth]{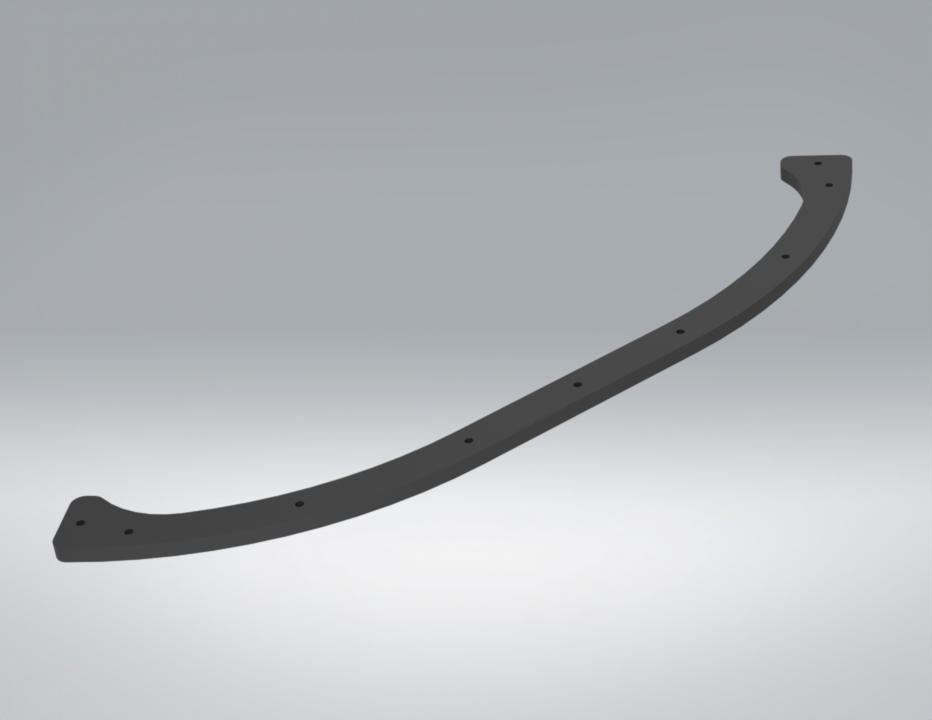}
    \caption{Middle Thruster Guard}
    \label{fig:middle_guard}
\end{figure}

\begin{figure}[h]
    \centering
    \includegraphics[width=0.6\linewidth]{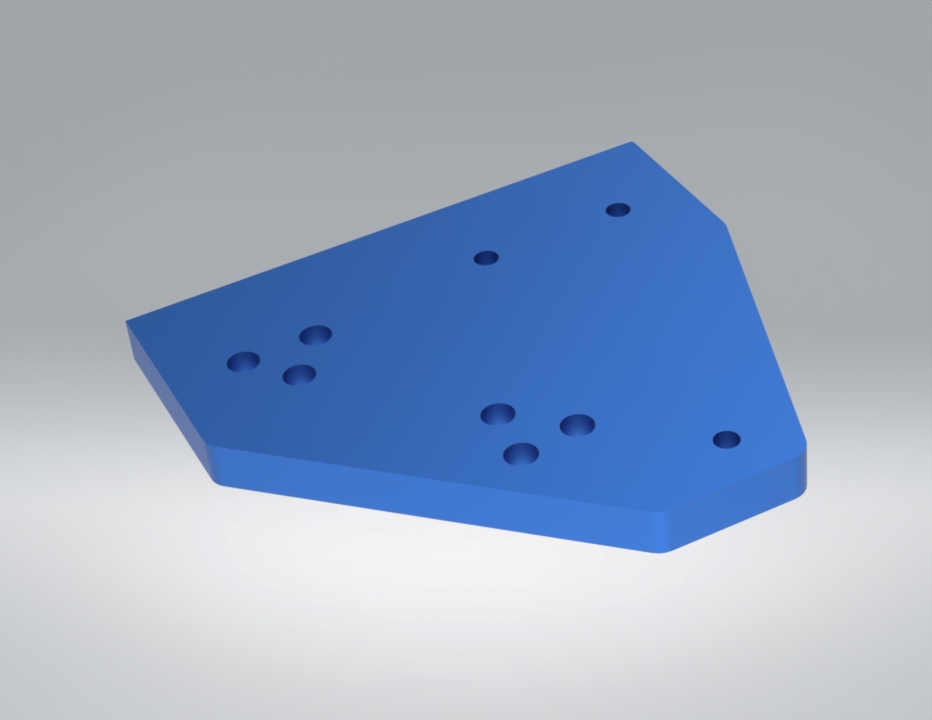}
    \caption{Diagonal Thruster Bracket}
    \label{fig:diagonal_bracket}
\end{figure}

\subsection{Custom Buoyancy Blocks}
We designed a new set of buoyancy mounts for Crush using SolidWorks CAD and Autodesk Fusion CAM. The blocks were machined on a 3-axis CNC mill. The shapes all match the inner profile of the case to fit as snug as possible and structurally support the case.

\begin{figure}[h]
    \centering
    \includegraphics[width=0.5\linewidth]{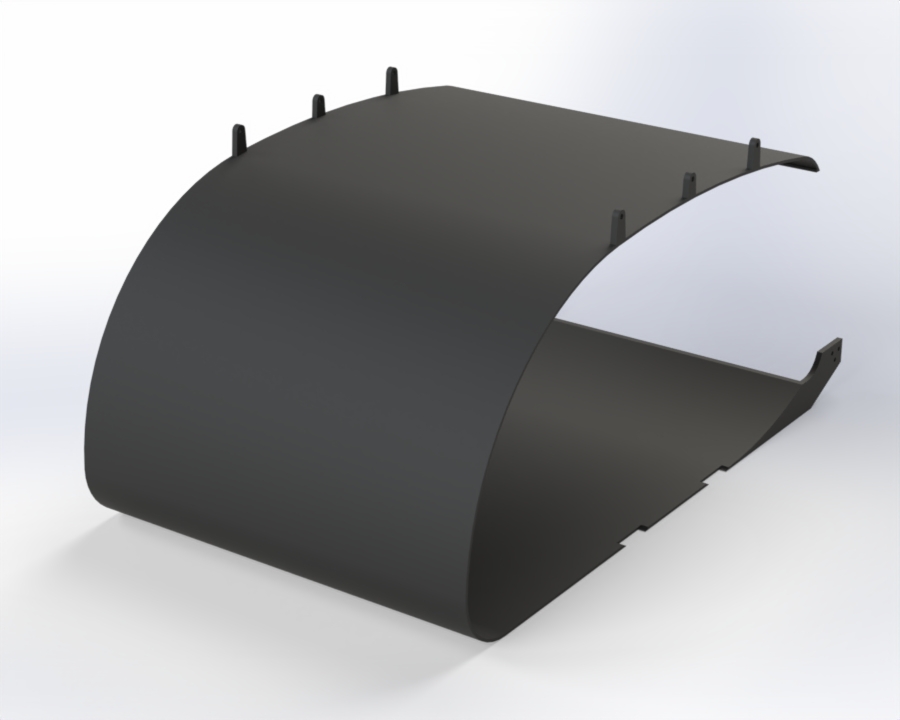}
    \caption{Crush's Case}
    \label{fig:case}
\end{figure}

\begin{figure}[h]
    \centering
    \includegraphics[width=0.5\linewidth]{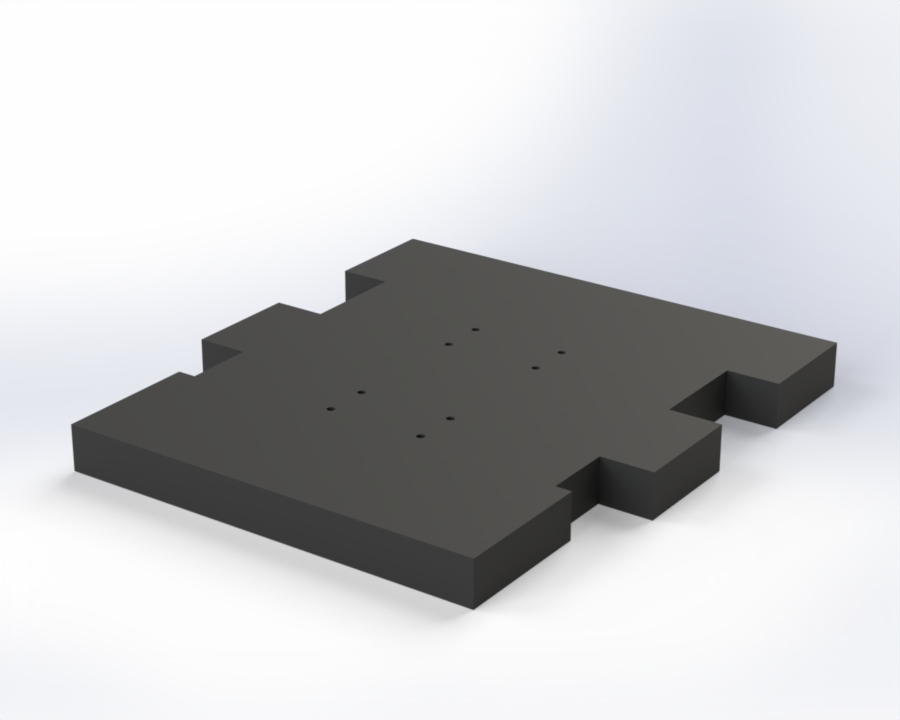}
    \caption{Crush's Bottom Buoyancy Block}
\end{figure}

\begin{figure}[h]
    \centering
    \includegraphics[width=0.5\linewidth]{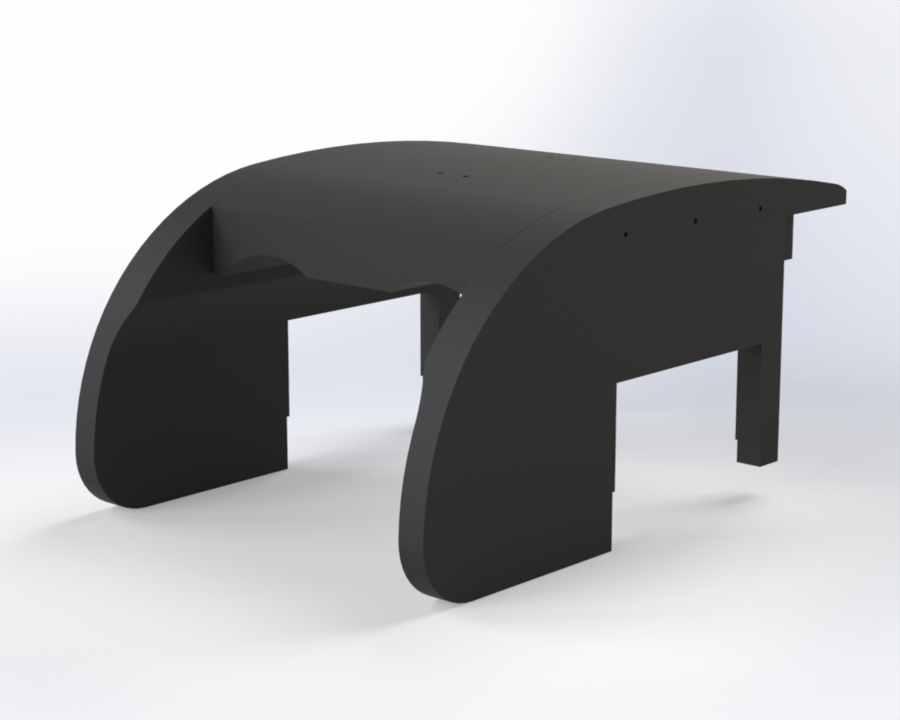}
    \caption{Crush's Top and side Buoyancy blocks}
\end{figure}

\subsection{Gripper/Claw}
The current gripper is mostly 3D printed and servo controlled with TPU claws to conform around any object. As the current four bar system leaves the gripper large and clunky, the next iteration will use compact parallel claws controlled via rack and pinion (Fig. 13).

\begin{figure}[h!]
    \centering
    \includegraphics[width=0.5\linewidth]{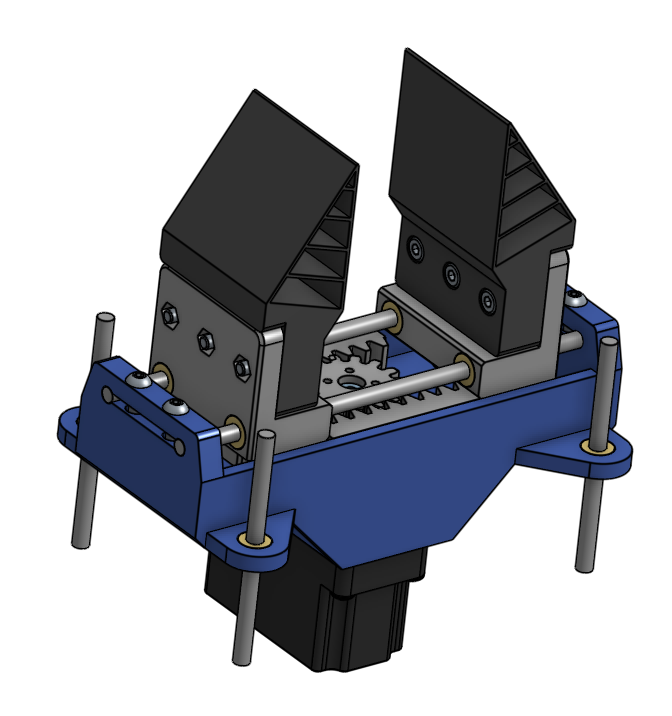}
    \caption{Parallel Axis Gripper}
    \label{fig:gripper}
\end{figure}

\subsection{Battery Capsule}
Frequent power disconnects meant we had to waste pool-testing time power cycling the robot. So, to increase reliability of the connection, we replaced the power wire and epoxied worn down connections. Then, we redesigned the battery holder to be more compact and reduce power cable strain. 

\subsection{Oogway Battery Capsule Clamps}
The previous clamps had large tolerances for low quality 3D printers, bad hinge geometry, and no access to the inner screw terminals. This year, we tightened tolerances, remodeled the joints, added access holes, and printed them in nylon 12 for its durability and inherent waterproofing.

\section{PCB-Based Electrical Stack}
\subsection{Electrical System Redesign for RoboSub 2027}

Looking ahead to \textbf{RoboSub 2027}, our biggest electrical goal is to redesign the current stack into a more integrated PCB-based system. Last year's design made important improvements in organization, especially through separated power domains, custom ESC boards, and robot-agnostic communication. However, the current implementation still relies on many jumper wires, screw terminals, and individual breakout boards. While these components are useful for rapid prototyping, they make the system harder to assemble consistently and can slow down debugging when something fails.

Our planned redesign will consolidate many of these functions into a unified stack of printed circuit boards. Instead of treating each board or module as a separate component connected by hand-routed wires, the new architecture will use PCBs to define the core electrical pathways of the robot, drastically reducing the prevalence of EMI. This will reduce wiring clutter, improve mechanical stability, and make the system easier to inspect and reproduce. We expect this redesign to reduce the size of the electrical stack
 by approximately \textbf{20\%}, which will free up valuable space inside the robot and make future maintenance easier.

\begin{figure}[h]
    \centering
    \includegraphics[width=1.0\linewidth]{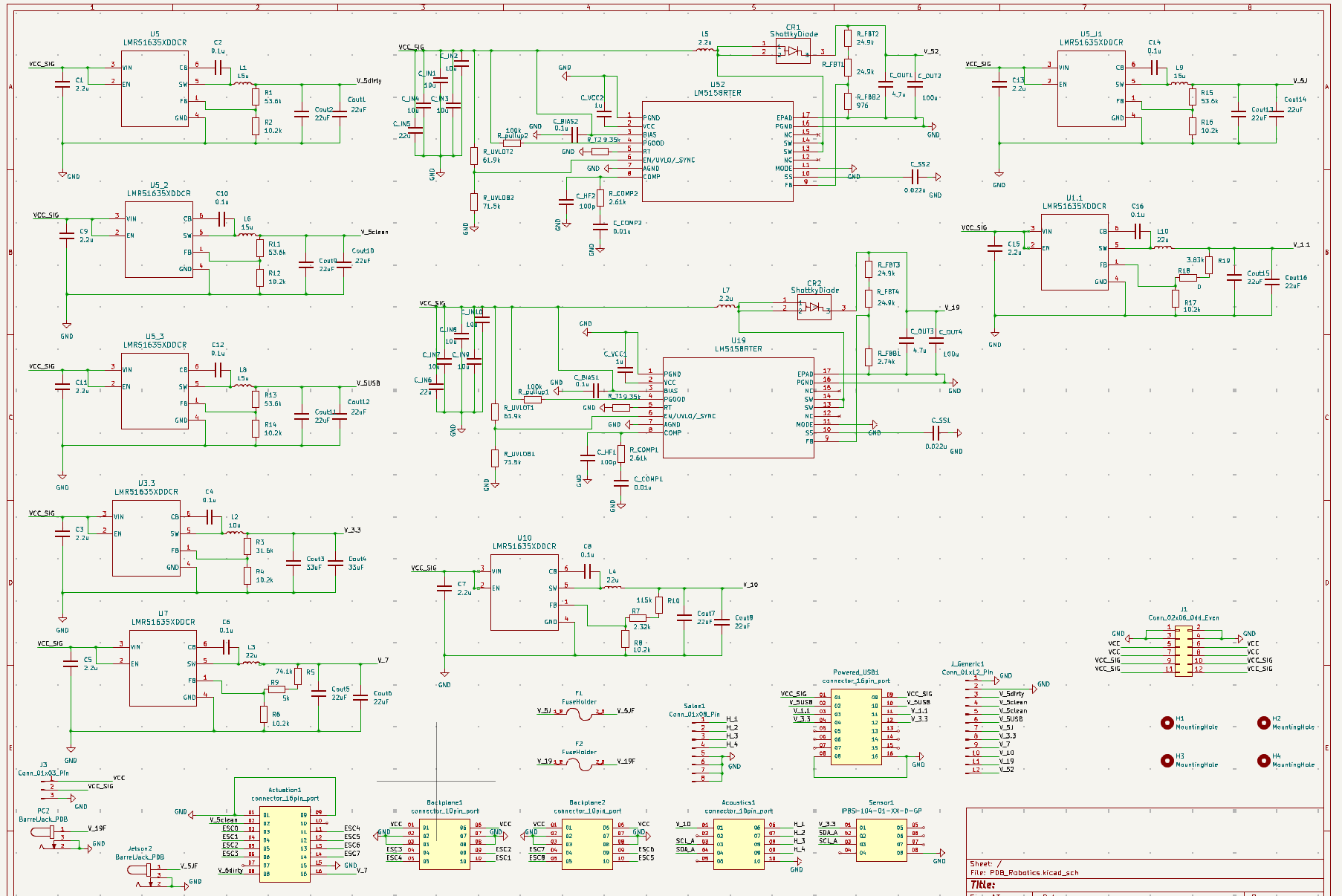}
    \caption{Power Distribution Board Schematic}
\end{figure}

\subsection{PCB-Based Modularity}

A major goal of the new PCB stack is to keep the flexibility of our current system while making it more robust. Each board will be designed around a specific electrical function, such as power distribution, sensor routing, actuator control, or communication. This will allow individual boards to be tested, replaced, or upgraded without requiring the entire electrical system to be rebuilt.

\begin{figure}[h]
    \centering
    \includegraphics[width=0.7\linewidth]{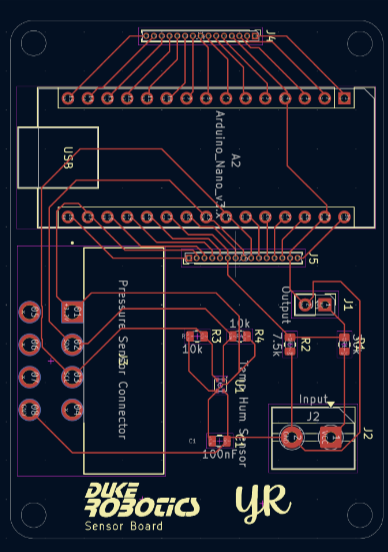}
    \caption{Sensor Board Physical Layout}
\end{figure}

This approach builds naturally from last year's custom PCB work. Last year, custom boards helped simplify ESC organization and power routing for Crush. For the 2027 redesign, we want to apply that same idea more broadly across the entire electrical stack. Instead of using PCBs for only select subsystems, the goal is to make the PCB stack the central structure of the electrical system. This will make the robots more plug-and-play and allow future team members to understand the system more quickly.

\subsection{Debugging and Maintainability}

One of the most important benefits of the new design will be easier debugging. In the current system, electrical problems can be difficult to isolate because many connections are physically similar, routed through tight spaces, or hidden underneath other components. When a subsystem fails, the team often has to determine whether the issue is caused by the component, the power source, the communication line, the firmware, or a loose connection. This can take significant time during pool testing, when debugging time is especially limited.

\begin{figure}[h]
    \centering
    \includegraphics[width=0.8\linewidth]{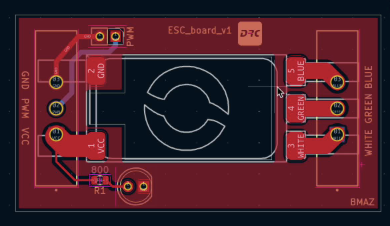}
    \caption{ESC Holder PCB Physical Layout}
\end{figure}

A PCB-based stack will make these problems easier to diagnose by giving each subsystem a more defined physical and electrical interface. Connections will be embedded into the boards rather than manually routed through large bundles of wires. We can also design the boards with clearer labeling, test points, status LEDs, and standardized connectors. These features will allow the team to check power rails, communication lines, and signal paths more quickly when something goes wrong.

\begin{figure}[h]
    \centering
    \includegraphics[width=0.9\linewidth]{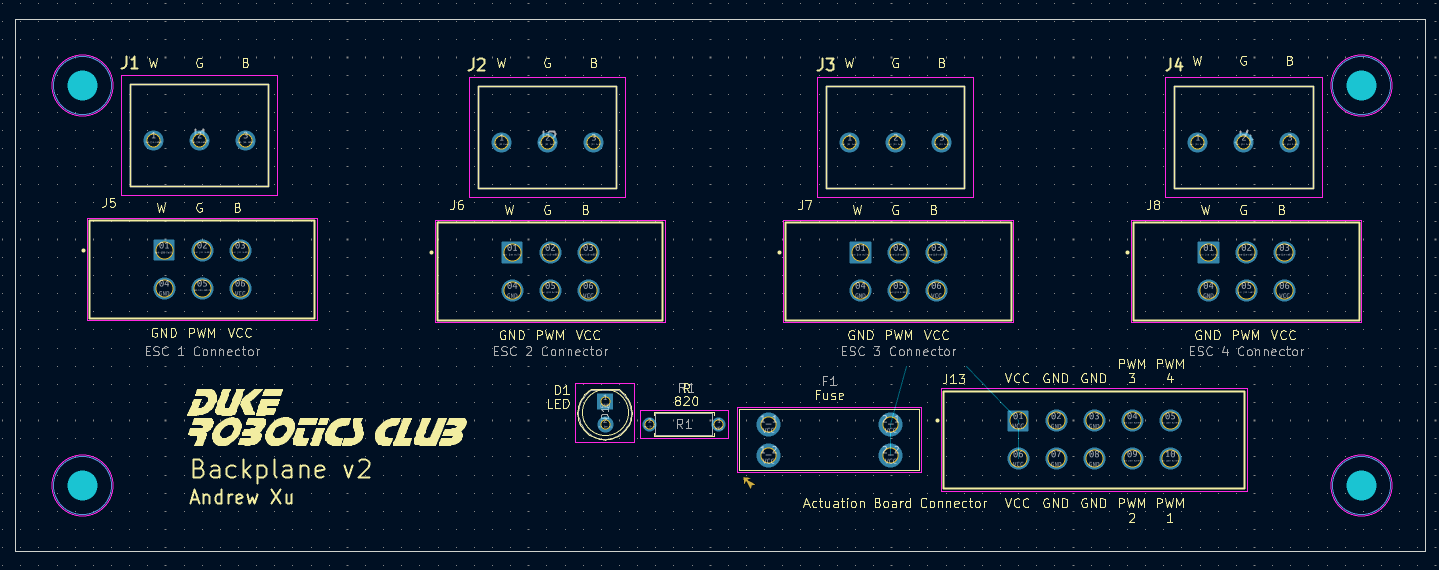}
    \caption{Backplane Physical Layout}
\end{figure}

This is especially important for long-term team sustainability. As new members join the electrical team, the system should be understandable without requiring extensive knowledge of every individual wire route. A cleaner PCB-based architecture will make onboarding easier and reduce the chance of wiring mistakes during maintenance or upgrades.

\section{Software}
\subsection{Jetson Nano YOLO Integration}
We began exploring future improvements to the CV pipeline by running YOLO directly on the Jetson Nano's GPU. Our current DepthAI-based pipeline has been useful, but depending entirely on DepthAI hardware and associated tooling limits flexibility and increases points of failure. Running a YOLO model directly on the Jetson would allow the robot to subscribe to a camera feed, perform inference onboard, and publish bounding boxes through ROS without relying on Luxonis-specific software processes, enabling further customization of our inference pipeline. 

Benchmarking the Jetson Nano provides a practical understanding of how its Maxwell‑based 128‑core GPU and quad‑core ARM CPU behave under real computer vision workloads. Because the Nano lacks Tensor Cores and relies on FP16/FP32 CUDA pipelines, performance varies significantly depending on model size, framework choice, and whether TensorRT optimizations are applied. Baseline PyTorch inference on models like YOLOv5n or YOLOv8n typically lands in the 3–7 FPS range. However, by activating CUDA, the inference drastically speeds up to around 30 FPS, and additional TensorRT‑optimized engines can push that into the 50 FPS range depending on input resolution and preprocessing overhead. CPU‑only execution, by contrast, is an order of magnitude slower and quickly saturates all four cores, making GPU acceleration essential for any real‑time or near‑real‑time application.

Due to the Or$\overset{\dotsaag}{\text{\i}}$n Nano's heavy graphical capabilities and conversely limited CPU, it was important to run this model using CUDA. CPU-only models would report around 300-400\% (or around 3-4 cores fully utilized), which was simply not feasible for running computer vision tasks in addition to managing the rest of the sensor stack. CUDA enabled the CPU to only be around 70-80\% utilized, varying based on the input resolution and computer vision task.

\begin{figure}[h]
    \centering
    \includegraphics[width=0.8\linewidth]{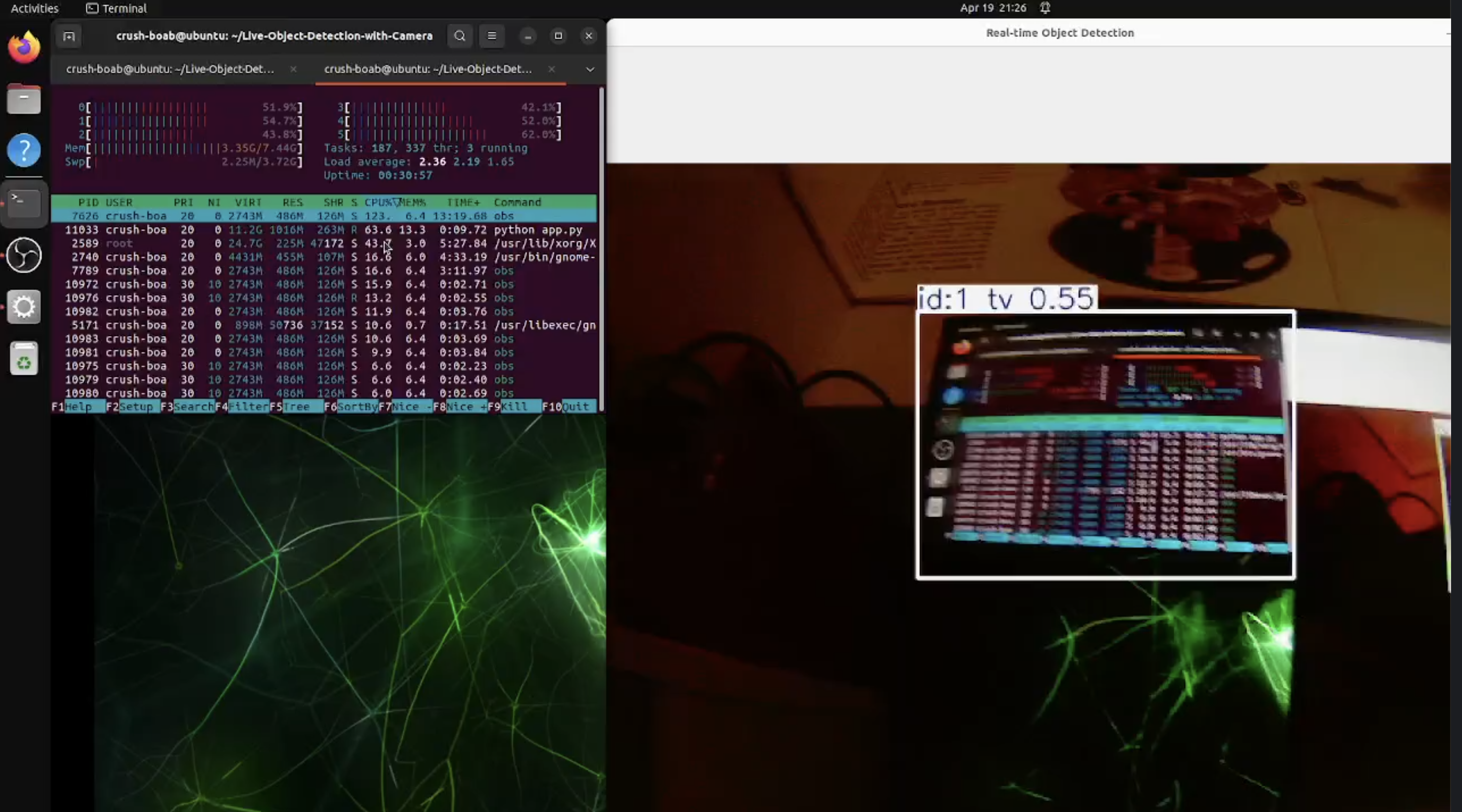}
    \caption{Jetson Nano Benchmarking}
\end{figure}

This work is still forward-looking, but it represents an important step toward a more flexible and maintainable perception stack.

\end{document}